\documentclass[lettersize,journal]{IEEEtran}
\usepackage{amsmath,amsfonts}
\usepackage{algorithmic}
\usepackage{algorithm}
\usepackage{array}
\usepackage{textcomp}
\usepackage{stfloats}
\usepackage{url}
\usepackage{verbatim}
\usepackage{graphicx}
\usepackage{subfigure}
\usepackage{multirow}
\usepackage{cite}
\usepackage{hyperref}
\usepackage{color}
\usepackage{ulem}

\begin{document}
\title{Multi-feature Reconstruction Network using Crossed-mask Restoration for Unsupervised Industrial Anomaly Detection}

\author{Junpu Wang, Guili Xu, Chunlei Li, Guangshuai Gao, Yuehua Cheng and Bing Lu

	\thanks{
        This work was supported in part by the National Natural Science Foundation of China under Grant 62472463, 62073161, 61905112.
        
		Junpu Wang, Chunlei Li and Guangshuai Gao are with School of Information and Communication Engineering, Zhongyuan University of Technology, Zhengzhou 450007, China. 
		
		Junpu Wang, Guili Xu$^{*}$ and Yuehua Cheng are with College of Automation Engineering, Nanjing University of Aeronautics and Astronautics, Nanjing 211106, China. $^{*}$ Corresponding author: Guili Xu (guilixu@nuaa.edu.cn)
		
		Bing Lu is with Applied Technology Research Institute, Henan Institute of Economics and Trade, Zhengzhou 450007, China. 
	}
}



\maketitle

\begin{abstract}
Unsupervised anomaly detection using only normal samples is of great significance for quality inspection in industrial manufacturing. Although existing reconstruction-based methods have achieved promising results, they still face two problems: poor distinguishable information in image reconstruction and well abnormal regeneration caused by model under-regularization. To overcome the above issues, we convert the image reconstruction into a combination of parallel feature restorations and propose a multi-feature reconstruction network, MFRNet, using crossed-mask restoration in this paper. Specifically, a multi-scale feature aggregator is first developed to generate more discriminative hierarchical representations of the input images from a pre-trained model. Subsequently, a crossed-mask generator is adopted to randomly cover the extracted feature map, followed by a restoration network based on the transformer structure for high-quality repair of the missing regions. Finally, a hybrid loss is equipped to guide model training and anomaly estimation, which gives consideration to both the pixel and structural similarity. Extensive experiments show that our method is highly competitive with or significantly outperforms other state-of-the-arts on four public available datasets and one self-made dataset.
\end{abstract}

\begin{IEEEkeywords}
Unsupervised anomaly detection, multi-feature reconstruction, crossed-mask restoration, transformer structure.
\end{IEEEkeywords}

\section{Introduction}
\IEEEPARstart{A}{nomaly} detection aims to identify and locate anomalous regions in images \cite{ref1}, which is a fundamental yet challenging task in visual understanding that has a wide range of real-world applications, such as product quality monitoring \cite{ref2,ref3,ref4}, medical health diagnosis \cite{ref5,ref6}, video surveillance \cite{ref7,ref8}, and so on. In the past decades, machine vision-based methods \cite{ref9,ref10} have gained significant attention, and are increasingly replacing the traditional manual inspection, which is time-consuming, subjective and inaccurate. However, these previous works generally rely on the extraction of handcrafted features based on the characteristics of a specific situation, limiting the generalization ability of the algorithms. Since the unprecedented breakthrough of deep learning in computer vision, great efforts have been made to apply it to high-quality anomaly detection. Nevertheless, in most real application scenarios, it is easy to acquire normal examples, but unrealistic to collect and label all types of abnormality, which makes it impossible to apply well-established full-supervised learning methods \cite{ref11,ref12}. Therefore, unsupervised learning methods that only require normal data for training are attracting more and more attention. The two classical paradigms for unsupervised anomaly detection methods are feature-based and reconstruction-based.

\begin{figure}
	\centering
	\includegraphics[width=0.5\textwidth]{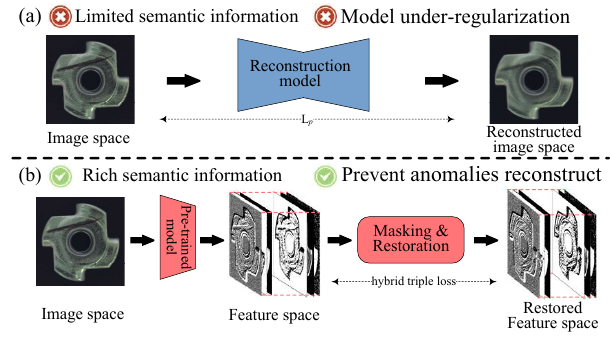}\\
	\caption{(a) Reconstruction model on image space tends to yield good reconstruction for anomalies using limited semantic information. (b) Our method reconstructs multi-scale features through a masking and restoration model for accurate anomaly detection.}\label{fig1}
\end{figure}

The former firstly extracts discriminative features for normal images using deep neural networks, including pre-trained models and self-supervised learning. Then, relevant statistical approaches, such as clustering algorithms or Gaussian models, are utilized to model the distribution of these normal features. During inference, examples that deviate from the learned distribution will be designated as anomalies. Since working on the feature space with a semantically meaningful representation, feature-based methods typically produce promising results. However, they usually lack interpretability, because it is unable to directly determine which part of the image causes a high abnormal score. Some works \cite{ref13,ref14} attempt to obtain anomaly localization by performing evaluations on a large number of image patches respectively through the sliding window strategy, which brings high complexity and limits the practical applications of these methods. Meanwhile, the artificially selected distribution assumption is hard to hold in all abnormal cases, ranging from subtle texture changes (e.g., weak scratches) to larger structural changes (e.g., missing components). This hinders the universal application of feature-based methods.

Another typical group of reconstruction-based methods commonly uses deep reconstruction models, such as auto-encoder (AE) and generative adversarial network (GAN), to model the latent representation of the normal data and then reconstruct itself. Since the trained models only have the knowledge specific to the normal samples, when the anomalous sample is fed through this pipeline, non-defective regions in it could be reconstructed reliably, while defective regions will most likely be poorly reconstructed. As a result, the difference between the input and its reconstruction is calculated by some distance metrics to generate an anomaly map, where the high reconstruction errors are designated as anomalies. Although reconstruction-based methods are intuitive and interpretable, their performance is relatively limited due to the several problems demonstrated in Fig.\ref{fig1} (a).

Firstly, most reconstruction-based models often fail to effectively utilize or learn semantic representation, which leads to imperfect anomaly detection in complex datasets. One main reason for this problem is that, unlike feature-based methods that inspect anomalies in feature space, most reconstruction-based models detect anomalies in image space with limited distinguishable information. Meanwhile, a recent study \cite{ref15} has shown that reconstruction-based models, especially vanilla AE, are likely to capture only low-level features rather than high-level semantic information, which is attributed to the information equivalence among the input and the reestablish target.

Secondly, the under-regularization ability of the reconstruction model makes it challenging to ensure a poor reconstruction of abnormalities. More specifically, reconstruction-based methods assume that reconstruction only succeeds when samples are normal. But due to the strong generalization ability of neural networks, anomalies could be regenerated well unexpectedly in practice, thus resulting in false detection. Several empirical methodologies \cite{ref7,ref16} have been used to address this problem, but the effectiveness remains limited. Consequently, exploring an effective regularization method to prevent the reconstruction model from regenerating anomalies remains an open question.

Thirdly, commonly used $L_p$ loss is a simple per-pixel distance metric, which ignores the inter-dependencies between the local image regions. As a consequence, these pixel-level measures will get high reconstruction errors even with slight localization imprecisions of edges, resulting in many false anomaly detection.

To address the above problems, we propose a universal anomaly detection architecture that combines the advantages of feature-based methods with reconstruction-based methods, named MFRNet. As shown in Fig.\ref{fig1} (b), to better distinguish the normal samples and abnormal samples, we perform reconstruction at the feature space extracted by a frozen pre-trained backbone. The generated multi-scale features have inherently multi-scale context-aware representations, which is beneficial for detecting anomalies with various scales. In addition, we break the information equivalence in traditional auto-encoders and convert anomaly detection into a restoration problem through crossed-masking followed by restoring. The masked regions are regenerated by conditional only on their surrounding context, which can effectively restrict the model’s capability to reconstruct anomalies. Finally, model training and anomaly estimation are performed using a hybrid loss to take into consideration both pixel and structural similarity.

Compared to the previous work, especially the preliminary reconstruction-based models, the contributions of this paper are summarized as follows.

1. We evaluate the difference between normal and abnormal images from the perspective of feature domain and propose an anomaly detection framework based on multi-scale feature reconstruction.

2. We consider the reconstruction task as a restoration problem to break information equivalence, which can learn more distinguishable semantic information and regularize the reconstruction model to prevent abnormal reconstruction.

3. A self-collected Fabric-US dataset containing 180 normal images for training and 400 abnormal images for testing is produced and publicly available. Extensive experiments carried out on Fabric-US and four other public datasets demonstrate the effectiveness and generalization of our method.

The rest of the paper is structured as follows. Section II reviews the related works on anomaly detection briefly, and Section III describes the framework of our proposed method in detail. Next, the experiment results on five datasets and an ablation study are reported in section IV. Then, section V provides some discussions about the proposed method. Finally, we summarize the work and present the future work in Section VI.

\section{Related work}
In this section, we restrict ourselves to an overview of current unsupervised anomaly detection methods for image data, focusing on those based on deep learning. They can be broadly divided into two categories: feature-based and reconstruction-based methods.

\subsection{Feature-based approaches}
One promising solution for unsupervised anomaly detection is to leverage relevant machine learning models to model the distribution of normal image features extracted by either handcrafted feature descriptors or deep neural networks. The images will be designated as anomalous if their corresponding features deviate from the learned distribution. Since the distribution modeling of normal samples and anomaly inference are both in the feature space, such methods are often referred as feature-based approaches. Obviously, the performance of this methodology is primarily influenced by two components: the feature extraction and distribution assumption modules. 

For the feature extraction modules, previous studies \cite{ref9,ref10} utilized a number of simple handcrafted feature descriptors, such as texture features or edge features, to represent the images. However, they generally fail to yield impressive detection performance, due to the lack of semantically discriminative features. With the rapid development of deep learning, in recent years, the trend has been shifted to use deep learning methods as feature extractors. SPADE \cite{ref17} could be the first work that directly used a deep neural network pre-trained on the large-scale image database ImageNet \cite{ref18} to obtain better feature representation. Subsequently, more and more researchers \cite{ref13,ref19,ref20,ref21,ref22} began to experiment with a variety of popular pre-trained networks, e.g., ResNet, Wide-ResNet or EfficientNet, and also produced decent anomaly detection results. Most of these studies discussed the effect of using different layer features on the experimental results, and declared that the ensemble of multiple feature representations of the networks can offer some benefits. However, there is still no general consensus on how to choose the appropriate feature layers for aggregation to cope with different scenarios. Besides using pre-trained deep networks, significant efforts had also been invested in designing proper self-supervised learning tasks to learn meaningful representations from scratch. The popular proxy tasks include predicting the relative position of a pair of random image patches \cite{ref23,ref24}, solving jigsaw puzzles of the randomly disrupted image patches \cite{ref25}, restoring the erased attributes of raw image (color and orientation information) \cite{ref15}, binary classification between the normal samples and the artificial pseudo-anomaly samples \cite{ref14,ref26}, obtaining the learned latent features from deep auto-encoders via CNN \cite{ref27,ref28} or vision transformer \cite{ref29}, and so on. Although these elaborate auxiliary tasks have shown great potential in extracting semantic features, the learned representation capacity will be greatly limited by the limited normal training data. Furthermore, such methods require an additional training phase, which is not off-the-shelf like pre-trained features, and they also don’t exhibit significant advantages over the ones using pre-trained features. Very recent works \cite{ref31,ref32} starting with \cite{ref30} are able to further transform the pre-trained features of normal data into a more simple well-defined distribution, typically Gaussian distribution, by using normalizing flow-based models. Due to being projected to a more tractable distribution, they generally showed favorable detection performance.

After obtaining meaningful feature representation describing normal data, various traditional statistical models are used to attempt to fit it into different hypothetical distributions. Then, the anomaly score can be calculated by the deviation distance between the features of test sample and the established distribution of normal features. Different feature-based methods mainly differ by the distribution estimation module that they adopt. The common distribution assumptions can be the k-nearest neighbours \cite{ref17,ref21}, support vector data description (SVDD) \cite{ref23,ref24}, Gaussian mixture model \cite{ref28,ref29}, multivariate Gaussian model \cite{ref13,ref14,ref19}, multiple independent multivariate Gaussian clustering \cite{ref3}, etc. Moreover, the seminal work \cite{ref33} and its successors \cite{ref34,ref34.5} used a student-teacher distillation scheme for anomaly detection and reached satisfactory results. Since the student networks are trained to regress the output of a teacher network pre-trained on ImageNet during the training, they only learn the normal data manifold and thus can be seen as an estimated distribution. Afterwards, the anomaly scores are derived from the discrepancy between the student network and teacher network. Although feature-based methods are extremely competitive, there are some crucial factors that affect their practical application in anomaly detection: (1) A heuristic distributional assumption is hard to characterize the distribution adequately as the data complexity increases. (2) Most of these methods lack interpretability that they simply predict the anomalies at the image-level without spatial localization. While performing anomaly evaluation for each image patch step-by-step can yield coarse detection results, it is computationally expensive. Remarkably, all of the above approaches indicate that normal and abnormal images are more likely to be distinguishable in feature space. This finding motivates our work to detect anomalies through the reconstruction of pre-trained deep features rather than raw images.

\subsection{Reconstruction-based approaches}
Another mainstream for anomaly detection is the reconstruction-based approach, which typically uses a reconstruction model (AE or GAN, for instance) to encode the latent manifold of training data and then reconstruct itself from there. At the inference stage, anomalous regions cannot be reconstructed faithfully from the learned latent representation of anomaly-free data. Since it is intuitive and explainable, i.e., the anomaly map is directly generated by the difference between the test image and its reconstructed image, this kind of method has received substantial attention.

\begin{figure*}[t]
	\centering
	\includegraphics[width=1\textwidth]{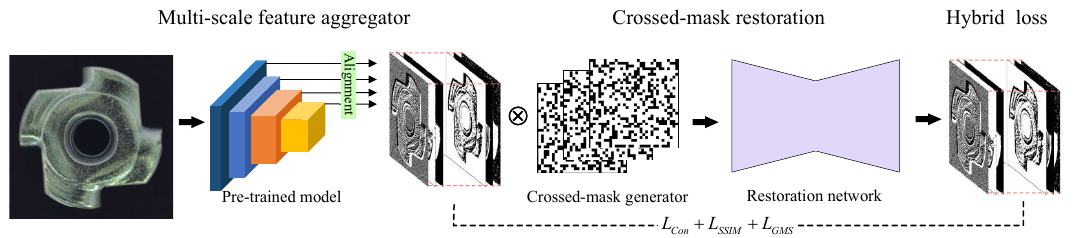}\\
	\caption{An overview of our unsupervised anomaly detection pipeline.}\label{fig2}
\end{figure*}

The most fundamental architecture of reconstruction-based methods, auto-encoder, is essentially an encoder-decoder network. The encoder manages to learn meaningful representations of normal inputs into a latent space, and the decoder attempts to regenerate the inputs from this low-dimensional space. The abnormal data unseen during the training are expected to not be reconstructed as accurately as normal ones, hence making higher reconstruction errors. Over the past few years, various variants of auto-encoder networks had been extensively explored for anomaly detection task. Chow et al. \cite{ref35} proposed to adopt convolutional AE for anomaly recognition, in which the hidden layers are implement by convolution. While Chen et al. \cite{ref36} implemented auto-encoder by the emerging Transformer structure to capture more non-local feature information. Further study \cite{ref37} attempted to inspect anomaly by the variational AE, but did not show significant performance gains over the convolutional AE. Additionally, the training of variational AE is relatively unstable. To improve the robustness of convolutional AE, Mei et al. \cite{ref38} introduced noise interference into the input images for model training and obtained a better detection performance. Although these models are supposed to only reconstruct normal images accurately, the under-regularization of neural networks sometimes result in anomalies being reproduced well, thus causing misdetection. Lots of efforts have been devoted to tackling this excessive generalization issue. \cite{ref7} believed that skip connections in most auto-encoders may simply copy the information of abnormal images, and thus proposed a U2-net-like backbone network without skip connections. Some works \cite{ref16,ref39} introduced a memory module into auto-encoder to record the normal patterns of the training samples. Given a defective input, the proposed memory-augmented AE will try to reproduce images using normal features stored in the memory module, reducing the possibility of recreating anomalies. To restrict the representation learning in the latent space explicitly, Liu et al. \cite{ref4} and Niu et al. \cite{ref40} proposed the novel auto-encoder with dual prototype loss, which were able to encourage the latent representation of encoder to keep closer to their own prototype, i.e., the center of the latent representation of training images. Very recently, Fei et al. \cite{ref15} held a view that the cause of anomaly reconstruction may be the failure to extract semantic features in the vanilla AE models. They suggested that the AE models tend to simply compress images instead of learning the abstract semantic representation, due to the information equivalence, namely the same information between the input and output. Later on, many following researchers \cite{ref41,ref42,ref43} adopted image inpainting for anomaly localization to break the information equivalence and thus obtained more semantic understanding of the image. In these methods, parts of the sample images are covered by pre-specified masks, and then these hide regions are restored conditioned on their surroundings by the inpainting models. During the testing, anomaly scores are determined by the discrepancy between the masked regions and the corresponding repairs. This effective paradigm also inspires our work to formulate the anomaly detection task as a restoration problem through cross-masking and then recovering. The vast majority of reconstruction-based approaches just perform reconstruction in raw image space, which is insufficient for complex scenarios. To introduce more image information into the reconstruction process, some attempts have been made to execute reconstruction in the space of pre-trained features \cite{ref36,ref44}, image pyramids \cite{ref38}, multi-frequency \cite{ref45,ref46} or texture–structure features \cite{ref47}. Meanwhile, there are some works trying to leverage different distance metrics to assess the difference between inputs and reconstructed outputs. A simple $L_1$ or  $L_2$ loss was used in \cite{ref35,ref36,ref44} for model training, however, this ignored the inter-dependencies between the neighboring regions. \cite{ref41,ref43} and \cite{ref48} introduced structural similarity losses to give a better measure for image similarity.

Another typical architecture is the generative adversarial network, which basically comprises a generator and a discriminator. Generator is to establish a mapping from random vectors drawn from a prior distribution to the input image space, while discriminator attempts to differentiate the created fake images by the generator from the real input images. Through a two-player zero-sum game \cite{ref49}, the trained network can learn the data distribution of input training data effectively, thereby producing samples as similar as possible to the normal samples. Initial effort in \cite{ref5} first applied GAN for anomaly detection. During the testing, the proposed AnoGAN is able to generate a normal image that is visually closest to the input image. And the anomaly map is obtained by comparing the discrepancy between the input image and the corresponding generated image. However, this model requires iterative backpropagation through the generator network to find the closest encoding to the defective image in prior distribution space, which is computationally expensive. To circumvent such problem, later improvement \cite{ref6} adopted an auto-encoder as the generator network to map the test image to a latent space and then output the normal version of the image. Moreover, the difference between each feature layer in the discriminator network can be regarded as a reliable criterion for identifying anomalies. Following this paradigm, most contemporary research \cite{ref7,ref40,ref45} added a discriminant network to the above-mentioned auto-encoders to improve the quality of the generated images by conducting adversarial training. Although some improvements have been made, generative adversarial networks tend to suffer from erratic training and therefore require constantly adjust the training period manually to obtain acceptable results.

\section{Methodology}

The problem definition of unsupervised anomaly detection is presented as follows: given a training dataset consisting only of normal images, the goal is to design a model that can determine whether an unseen test image is defective or not and further locate anomalous regions in the defective images. To solve this problem effectively, we develop a multi-feature reconstruction network using crossed-mask restoration. As shown in Fig.\ref{fig2}, it is mainly composed of three parts: (1) a multi-scale feature aggregator to extract discriminative hierarchical features of the raw image; (2) a crossed-mask restoration network trained over the generated multi-scale representation to partially cover and recover it again; (3) a hybrid loss to measure the discrepancy between the input and regeneration more comprehensively. Since our model is trained only with normal images, the feature restoration of abnormal regions will fail, and the restoration degree can be used for anomaly inference. In the following subsections, our anomaly detection framework will be described elaborately.

\subsection{Multi-scale feature aggregator}
It has been certified that feature representation from a pre-trained model could provide more distinguishable information for identifying normal and abnormal images \cite{ref33}. Therefore, a frozen pre-trained model is firstly adopted to extract multi-scale features in this paper. Take VGG16 \cite{ref50} as an example, we can obtain a set of feature maps with different sizes $\left\{ {f_1 (I),f_2 (I), \ldots ,f_l (I)} \right\}$ for a raw image $I$, where $f_l (\cdot)$ means the output of $l$-th convolutional layer. To achieve effective alignment, these feature maps are resized to the same size, and concatenated along the channel axis to produce a multi-scale feature map:
\begin{equation}
F(I) = {\mathop{ cat}\nolimits} \left( {re(f_1 (I)),re(f_{\rm{2}} (I)), \ldots ,re(f_l (I))} \right) \in \mathbb{R}^{W \times H \times C} 
\end{equation}
where $cat$ and $re$ denote the operations of concatenation and resize, respectively. $H$, $W$ and $C$ are height, width and channel dimensions of the produced feature map. 

Obviously, feature maps from different convolutional layers contain different information and thus are sensitive to the different abnormalities. Earlier layers output higher-resolution maps encoding more low-level information (e.g., texture and edge), while latter layers output lower-resolution maps encoding richer semantic information. As a result, the aggregation of these complementary multi-scale features plays an important role in anomaly detection.

\subsection{Crossed-mask restoration network}
After the multi-scale features of the image are ready, we leverage a crossed-mask restoration network to extend the reconstruction-based approach for anomaly localization. Specifically, a set of complementary crossed masks are generated to formulate a restoration task, followed by a restoration network trained to predict the covered regions. In this way, the difference between the masking region and the corresponding restoration is going to be significant for anomalous regions. By covering and restoring the partial regions of the input, our method can gain a much deeper semantic understanding of the scene, and regularizes the reconstruction model to prevent anomalies from being reconstructed well.

\subsubsection{Crossed-mask generator}
\begin{figure}[t]
	\centering
	\includegraphics[width=0.49\textwidth]{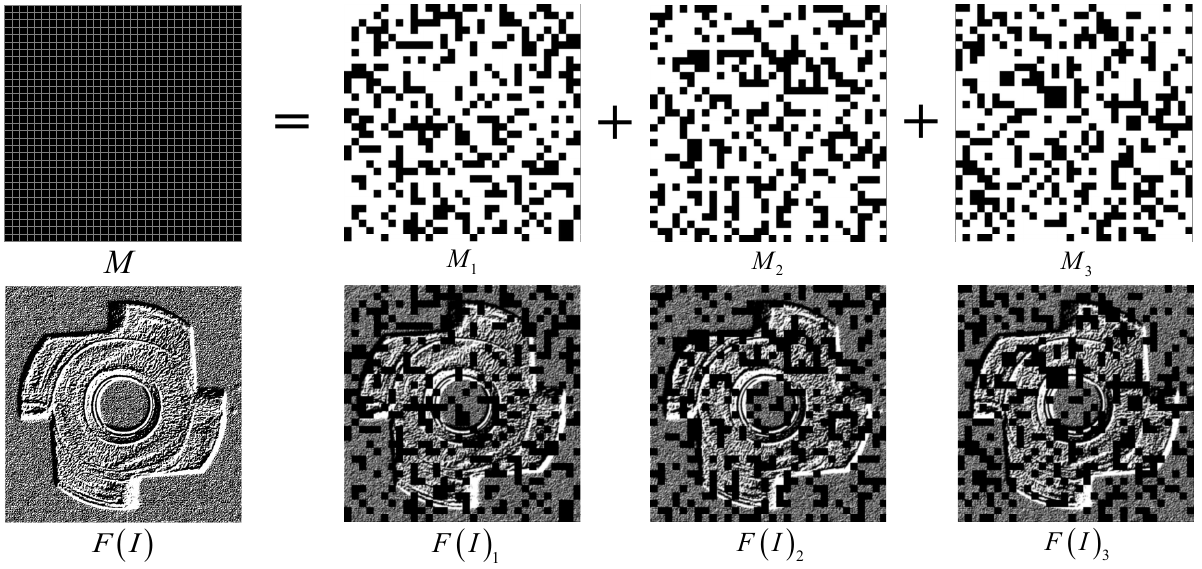}\\
	\caption{Example of masking operation with parameters $k=2$ and $n=3$ on a single-channel feature map. The masked regions within each subset are complementary, thus covering all possible anomalies.}\label{fig3}
\end{figure}

An all-zero mask $M$ with the same size as the multi-scale feature map is uniformly partitioned into a grid of square regions of the size $k \times k$. And then these $\frac{H}{k} \times \frac{W}{k}$ grids are randomly divided into $n$ disjoint sub-masks $M_i \left( {i = 1, \ldots n} \right)$. Therefore, we have $M = \bigcup\nolimits_{i = 1}^n {M_i } $. These complementary sub-masks, after being replicated along the channel dimension, are respectively multiplied by the multi-scale feature map to obtain $F\left( I \right)_i  = F\left( I \right) \times M_i $, which serves as the input for the later restoration network. Since these complementary sub-masks can jointly cover all the pixels of the feature map, any possible anomalous regions will not be lost. For convenience, we illustrate the masking operation with parameters $k=2$ and $n=3$ on a single-channel feature map in Fig.\ref{fig3}.

Since the sizes of anomalies are diverse, multiple restorations using different $k$ values sampled from a set $K$ should be considered to obtain more reliable detection results. Meanwhile, the parameter $n$ could influence the representation capacity and repairing quality of restoration network, thus determining the performance of anomaly detection. Therefore, the proper match of parameters $K$ and $n$ is of great importance for our method, which will be discussed in Section ‘Discussion’.

\subsubsection{Restoration network}
\begin{figure}[h]
	\centering
	\includegraphics[width=0.48\textwidth]{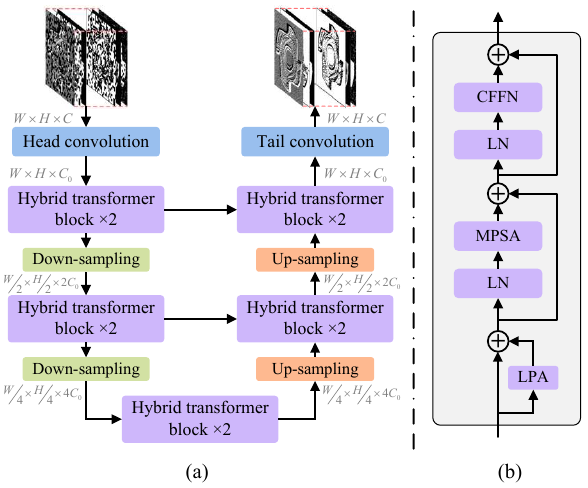}\\
	\caption{(a) Architecture of the proposed restoration network. (b) Stucture of the hybrid transformer block.}\label{fig4}
\end{figure}

Given a masked feature map $F\left( I \right)_i \in \mathbb{R}^{W \times H \times C} $, we employ a concise yet powerful restoration network to imagine visually reasonable and semantically consistent content for the missing regions. As shown in Fig.\ref{fig4} (a), it overall follows the principle of the conventional auto-encoder with skip connections, and is mainly composed of a head convolution, a tail convolution and a hybrid transformer body, combining the merits of CNN and recently emerging transformer structures \cite{ref51}. 

The head convolution is used to reduce the input feature dimension to $C_0 $ using a convolutional layer with the kernel size of 3$\times$3. The benefits are twofold: On the one hand, this dimensionality reduction operation can greatly reduce computational complexity and memory overhead. On the other hand, it can facilitate the information interaction across channels, resulting in a richer representation capability. The tail convolution is structurally identical to the head convolution, also a 3$\times$3 convolution, and is located behind the hybrid transformer body. It is designed to modulate the feature dimension to match the original input for anomaly estimation. The hybrid transformer body is comprised of five stages, each of which involves a stack of hybrid transformer blocks. Furthermore, a down-sampling layer is used to halve the size and double the channel of the input map in the first two stages, while an up-sampling layer is used to do the opposite in the last two stages. Both of these are implemented by convolutions. Fig.\ref{fig4} (a) shows the corresponding feature sizes and the number of hybrid transformer blocks in each stage. The details of the hybrid transformer block are presented below.

The hybrid transformer block is a light-weight transformer variant, which is illustrated in Fig.\ref{fig4}(b). We abandon the complex absolute or relative positional encodings used in previous transformer architectures and adopt a more convenient and efficient locally position-aware (LPA) module. Structurally, LPA is a 3$\times$3 convolution with zero padding 1, which can implicitly provide local spatial relationships for our hybrid transformer block. Subsequently, Layer normalization (LN) and residual connection are applied at both ends of the multi-pooling self-attention (MPSA) and convolutional feed-forward network (CFFN) respectively. CFFN is made up of two 1$\times$1 convolutions and one 3$\times$3 convolution with activation functions in between for non-linear feature transformation. To lower the high computational cost of previous self-attention module, we apply a multi-pooling operation to the computation of key and value matrices in our MPSA. Specifically, suppose the input of our MPSA is $z$, then the output is formulated as:
\begin{equation}
{\mathop{\rm Softmax}\nolimits} \left( {\frac{{zw^q (z'w^k )^T }}{{\sqrt d }}} \right)\left( {z'w^v } \right)
\end{equation}
where $w^q $, $w^k $ and $w^v $ are the linear projections used to calculate the query, key, and value matrices. The notation $d$ is the dimension of $z'w^k $ for scale balance. $z'$ is a concatenation of $z$ after passing through multiple average pooling layers with a set of ratios $R$, i.e., $z' = cat\left( {\left. {{\mathop{\rm avgpool}\nolimits} _i \left( x \right)} \right|i \in R} \right)$. For sufficiently large pooling ratios, $z'$ can be a smaller substitute for $z$ and contain all of its contextual information.

The transformer structure in our restoration network can explicitly learn long-range dependencies to speculate the semantically plausible appearance, while the intrinsic locality of CNN makes it more adept at extracting local features to recover the texture details. Consequently, the high-quality reconstruction $\hat F\left( I \right)_i $ for each masked map $F\left( I \right)_i $ can be generated independently. And the final feature reconstruction result $\hat F\left( I \right)$ is a union of the newly restored regions in each partial reconstruction, i.e., $\hat F\left( I \right) = \bigcup\nolimits_{i = 1}^n {\hat F\left( I \right)_i  \times \left( {1 - M_i } \right)} $.

\subsection{Hybrid loss}
To enable our crossed-mask restoration network to reconstruct the input well, a hybrid loss is adopted in this paper. It combines three loss functions to take into consideration both the pixel and structural similarity.
\subsubsection{Contextual loss}
To explicitly force the network to generate contextually similar feature map, we apply a $L_2$-loss between the multi-scale feature map of an image $F\left( I \right)$ and it’s corresponding regenerated feature map $\hat F\left( I \right)$. It is expressed as:
\begin{equation}
L_{Con}  = \frac{1}{{H \times W}}\left\| {F\left( I \right) - \hat F\left( I \right)} \right\|_2^2 
\end{equation}

However, such per-pixel measure could ignore the interdependency between the neighboring regions, which is often unreasonable. Therefore, we additionally introduce two similarity metrics, including structural similarity index (SSIM) \cite{ref52} and gradient magnitude similarity (GMS) \cite{ref53} to perceive different structural properties.

\subsubsection{SSIM loss}
The SSIM loss is denoted as:
\begin{equation}
L_{SSIM}  = \frac{1}{{H \times W}}\sum\limits_{i = 1}^H {\sum\limits_{j = 1}^W {\left( {1 - SSIM(F\left( I \right),\hat F\left( I \right))} \right)_{\left( {i,j} \right)} } } 
\end{equation}
where 1 means a matrix of ones with the same size as $F\left( I \right)$. $SSIM(F\left( I \right),\hat F\left( I \right))$ is the structural similarity map between $F\left( I \right)$ and $\hat F\left( I \right)$. The value of each coordinate in the map is the $ssim$ value between two patches $x$ and $y$ cropped from $F\left( I \right)$ and $\hat F\left( I \right)$ centered at $\left( {i,j} \right)$, which can be denoted as:
\begin{equation}
ssim(x,y) = \frac{{\left( {2\mu _x \mu _y  + a_1 } \right)\left( {2\sigma _{xy}  + a_2 } \right)}}{{\left( {\mu _x^2  + \mu _y^2  + a_1 } \right)\left( {\sigma _x^2  + \sigma _y^2  + a_2 } \right)}}
\end{equation}
where $\mu _x $, $\mu _y $, $\sigma _x^2 $ and $\sigma _y^2 $ are the mean and variance of $x$ and $y$ respectively, and $\sigma _{xy} $ is the covariance of them. $a_1 $ and $a_2 $ mean two small positive constants to avoid dividing by zero.

\subsubsection{GMS loss}
The GMS loss is represented as:
\begin{equation}
L_{GMS}  = \frac{1}{{H \times W}}\sum\limits_{i = 1}^H {\sum\limits_{j = 1}^W {\left( {1 - GMS(F\left( I \right),\hat F\left( I \right))} \right)_{\left( {i,j} \right)} } } 
\end{equation}
where $GMS(F\left( I \right),\hat F\left( I \right))$ denotes the gradient magnitude similarity map between $F\left( I \right)$ and $\hat F\left( I \right)$, which is shown below:
\begin{equation}
GMS(F\left( I \right),\hat F\left( I \right)) = \frac{{2G\left( {F\left( I \right)} \right)G\left( {\hat F\left( I \right)} \right) + b}}{{G\left( {F\left( I \right)} \right)^2  + G\left( {\hat F\left( I \right)} \right)^2  + b}}
\end{equation}
where $b$ is small positive constant to avoid instability. $G\left( {F\left( I \right)} \right)$ is the gradient magnitude map for $F\left( I \right)$ that can be computed by $\sqrt {\left( {h_1  \otimes F\left( I \right)} \right)^2  + \left( {h_2  \otimes F\left( I \right)} \right)^2 } $, where $h_1 $ and $h_2 $ indicate two 3$\times$3 Prewitt filters along the horizontal and vertical directions and symbol $\otimes $ denotes the convolution operation.

Finally, the hybrid loss is a sum of the above losses:
\begin{equation}
L = L_{Con}  + L_{SSIM}  + L_{GMS} 
\end{equation}

\section{Experiment results}
The experimental settings, including evaluation metrics, datasets description, implementation details and competing methods, are first introduced. Then, the generalization ability and superiority of our method are demonstrated by analyzing the comparison results against state-of-the-art methods on four publicly-available datasets and our newly-built dataset. Finally, we also report the performance of our method with different components to illustrate the contribution of the individual designs.

\subsection{Experimental setups}
\subsubsection{Evaluation metrics}
Besides the visual comparison, we introduce several evaluation metrics for a more comprehensive and objective assessment. Following the recent work \cite{ref2} on anomaly detection, Area Under the Receiver Operating Characteristics (AUROC) is adopted as the primary indicator for performance evaluation. Meanwhile, to increase the diversity of evaluation standards, we also introduce three widely-used criterions in segmentation task, including MAE, ACC and F1-score. As the general rule, higher AUROC, ACC, F1 score, and lower MAE mean better anomaly detection performance. 

MAE represents the mean absolute pixel-wise difference between the predicted detection result $p$ and its ground truth $g$, formulated as:
\begin{equation}
{\rm{MAE}} = \frac{1}{{w \times h}}\sum\limits_{i = 1}^w {\sum\limits_{j = 1}^h {\left| {p\left( {i,j} \right) - g\left( {i,j} \right)} \right|} } 
\end{equation}
where $w$ and $h$ denote the width and height of image, respectively, and $\left( {i,j} \right)$ represents the image pixel coordinates.

ACC refers to the accuracy rate, which is computed by the ratio of the properly predicted pixels and the total pixels. As expressed in the following formula:
\begin{equation}
{\rm{ACC = }}\frac{{{\rm{TP + TN}}}}{{{\rm{TP + FP + TN + FN}}}}
\end{equation}
where TP, TN, FP and FN indicate the number of true positive (defective pixels correctly detected), false positive (normal pixels misclassified as defective), true negative (defect-free pixels successfully detected) and false negative (defective pixels misclassified as normal), respectively.

F1 score denotes an overall metric for performance evaluation, calculated by the harmonic mean of $precision$ and $recall$, as follows:
\begin{equation}
F1 = {\rm{2}}\frac{{precision \cdot recall}}{{precision{\rm{ + }}recall}} = \frac{{2\rm{TP}}}{{\rm{2TP + FP + FN}}}
\end{equation}
where $precision$ denoted as $\rm{TP}/\rm{(TP+FP)}$ is the fraction of defective pixels among the detected pixels. $recall$ expressed as $\rm{TP}/\rm{(TP+FN)}$ is the fraction of anomalous pixels that are correctly identified.

\subsubsection{Datasets description}
~\linebreak[1]

\textit{a) MVTec AD dataset \cite{ref2}:} It is the first comprehensive real-world industrial image dataset for anomaly detection, which is composed of 5,354 high-resolution images with a range from 700$\times$700 to 1024$\times$1024 pixels. There are five texture and ten object categories from different application scenarios. Each category consists of around 60 to 391 normal images for training, and the remaining normal images as well as anomaly images with various types of defect for testing. The relatively small number of training images poses a challenge for learning effective deep representations.

\textit{b) BTAD dataset \cite{ref29}:} The dataset is made up of three different types of industrial products, called product 01, product 02 and product 03. The training sets for the three categories include 400 (1600$\times$1600 pixels), 399 (600$\times$600 pixels) and 1000 (800$\times$600 pixels) defect-free images, respectively, while the testing set is a mixture of normal images and abnormal images. For each anomalous image, there is a detailed pixel-wise ground truth annotation. The high complexity of texture makes it challenging for anomaly detection.

\textit{c) MT dataset \cite{ref54}:} This dataset is a magnetic tile surface defect dataset, which involves 392 defective and 925 defect-free images with varied illuminations and image sizes. There are five defect types including blowhole, crack, fray, break and uneven occurring in the testing set. In our implementation, the 925 normal samples are served as the training set and 392 abnormal samples as the testing set. The existence of pseudo-defects and uneven lighting causes great difficulty in the detection of defects.

\textit{d) MSD-US dataset \cite{ref55}:} Such dataset is a recently released mobile phone screen surface defect dataset collected by an industrial camera. It contains of three types of man-made defects: oil, stain and scratch. Each class consists 400 images with 1920$\times$1080 pixels for testing, and the corresponding ground truth masks are given. To follow the unsupervised setting, 20 extra non-defective images are provided for training. Uneven illumination at the edge of screen brings great challenges to anomaly detection.

\begin{figure}[t]
	\centering
	\includegraphics[width=0.495\textwidth]{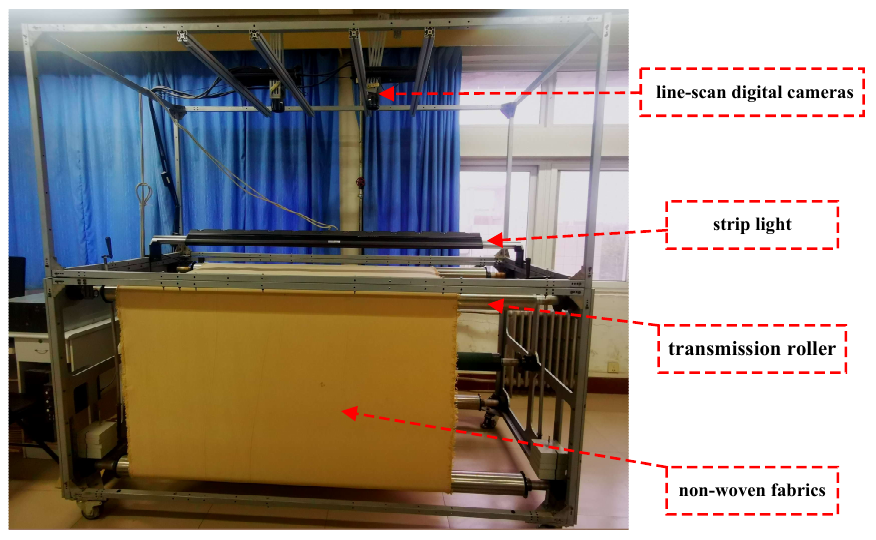}\\
	\caption{The proposed acquisition equipment for fabric image.}\label{fig5}
\end{figure}

\textit{e) Fabric-US dataset:} This is a novel dataset of fabric images collected by our proposed acquisition equipment shown in Fig.\ref{fig5}. The acquisition system primarily consists of line-scan digital cameras, strip lighting, and transmission roller. The line-scan cameras are equipped with high-precision CCD sensors, offering a lateral resolution of 4K and capable of achieving a maximum scanning speed of 16 meters per second. The strip light is strategically positioned perpendicular to the fabric surface, ensuring sufficient brightness to enhance the visibility of defects and make them more discernible. Due to the inherent softness and flexibility of fabric, the transmission roller facilitates the controlled movement and rolling of the material during the scanning process, ensuring stability and consistent image capture. This integrated system provides a reliable and efficient platform for high-quality fabric image acquisition, supporting detailed analysis and defect detection. Finally, 180 normal images are captured to serve as a training set, and 400 typical defect images are adopted as abnormal samples for the testing. The resolution of each raw image is 512$\times$512 pixels. Since the weak defect features, this dataset becomes challenging. This dataset is the unsupervised version of previous Fabric dataset \cite{ref56}, thus named Fabric-US (UnSupervised), and is available to the public\footnote{\href{https://pan.baidu.com/s/19gZoBxI9AUCQUTTBxyq8mA?pwd=0927}{https://pan.baidu.com/s/19gZoBxI9AUCQUTTBxyq8mA?pwd=0927}}.

\begin{table*}[]
	\centering
	\caption{F1 results of different methods on the MVTec AD dataset ($\%$).The top three performances are highlighted with \textbf{bold} and \underline{underline}}\label{table1}
	\setlength{\tabcolsep}{0.8mm}
	\renewcommand\arraystretch{1.5}
	\begin{tabular}{l|c|c|c|c|c|c|c|c|c|c|c|c|c|c|c|c}
		\hline
		Methods    & Bottle & Cable  & Capsule & Carpet & Grid  & Hazelnut & Leather & Metal\_nut & Pill  & Screw & Tile  & Toothbrush & Transistor & Wood  & Zipper & Mean  \\ \hline
		Patch-SVDD \cite{ref23} & 49.97	&25.86	&6.62	&22.77	&5.86	&22.69	&14.05	&41.58	&19.91&	1.66	&21.48	&29.78	&20.46	&17.92	&18.54	&21.28  \\ \hline
		SPADE \cite{ref17}      & 48.79	&35.20	&28.44	&46.83	&22.23	&38.22	&32.64	&41.15	&28.45	&17.65	&47.41	&23.30	&43.31	&39.99	&40.15	&35.58 \\ \hline
		GCPF \cite{ref3}       & 59.47	&\textbf{49.55}	&23.52	&49.10	&27.01	&52.2	&34.7&	42.37	&25.55	&11.83	&\textbf{55.27}	&16.63	&41.38	&42.27	&41.44	&38.15 \\ \hline
		AE-ssim \cite{ref48}    & 46.43	&18.46	&15.75	&11.71	&8.90	&49.54	&39.59	&40.1	&22.71	&26.79	&14.64	&39.09	&36.65	&16.31	&30.82	&27.83 \\ \hline
		GANomaly \cite{ref8}   & 13.08	&14.73	&15.66	&16.37	&16.52	&29.44	&28.47	&18.48	&19.93	&20.83	&20.83	&28.74	&11.20	&19.54	&5.42	&18.62 \\ \hline
		STPM \cite{ref57}       &\underline{62.41}	&49.41	&19.64	&49.31	&26.55	&41.07	&16.11	&44.98	&38.25	&16.34	&46.91	&26.76	&\textbf{49.18}	&38.39	&40.46	&37.72 \\ \hline
		MKDAD \cite{ref34}      &59.38	&27.27	&30.93	&53.21	&29.89	&39.29	&44.94	&50.37	&35.10	&12.95	&43.73	&34.42	&42.76	&\underline{46.02}	&37.38	&39.18 \\ \hline
		DifferNet \cite{ref30}  & 10.26	&9.69	&10.50	&15.92 &13.50	&15.84	&25.93	&11.32	&14.82	&10.01	&9.97	&16.74	&9.93	&10.66	&15.42	&13.37 \\ \hline
		CS-Flow \cite{ref32}    &44.30	&35.48	&21.79	&43.96	&24.89	&30.53	&34.98	&41.10	&25.13	&18.40	&46.68	&24.78	&38.35	&31.81	&43.23	&33.69 \\ \hline
		DFR \cite{ref44}        & 49.73	&40.64	&\underline{54.03}	&\underline{53.61}	&\underline{51.45}	&\textbf{59.31}	&\underline{60.87}	&\textbf{54.52}	&\textbf{47.85}	&\underline{40.61}	&41.82	&\underline{45.97}	&39.93	&44.90	&51.26	&\underline{49.10} \\ \hline
		OCR-GAN \cite{ref45}    & 8.94	&3.05	&11.02	&1.21	&1.81	&16.05	&8.56	&3.37	&4.72	&7.12	&4.51	&4.79	&2.32	&7.63	&5.17	&6.02 \\ \hline
		UTRAD \cite{ref36}      &31.32	&4.45	&14.94	&16.13	&10.82	&23.41	&37.54	&11.04	&13.58	&0.72	&21.99	&19.35	&6.35	&26.20	&18.56	&17.09 \\ \hline
		RIAD \cite{ref41}       &52.52	&33.64	&44.91	&37.37	&39.98	&52.24	&51.26	&44.14	&27.47	&31.48	&30.91	&40.10	&40.62	&36.36	&\underline{52.96}	&41.06 \\ \hline
		SSM \cite{ref43}        & 37.75	&25.15	&34.03	&17.35	&35.50	&54.23	&52.87	&30.73	&25.31	&31.8	&23.31	&33.34	&28.73	&34.83	&33.70	&33.24 \\ \hline
		Ours       & \textbf{64.01}	&\underline{49.50}	&\textbf{55.30}	&\textbf{57.21} &\textbf{52.65}	&\underline{57.21}	&\textbf{61.49}	&\underline{50.75}	&\underline{45.59}	&\textbf{43.36}	&\underline{48.32}	&\textbf{47.09}	&\underline{47.65}	&\textbf{47.50}	&\textbf{53.97}	&\textbf{52.11} \\ \hline
	\end{tabular}
\end{table*}

\subsubsection{Implementation details}
For the input image, its resolution is resized to 256$\times$256 uniformly, and its channel is triplicated when encountering a grayscale image. For the multi-scale feature aggregator, we choose a frozen pre-trained model, VGG16, as the feature extractor by default. Considering the computational cost, the outputs of its first three blocks are selected for the experiments, and the size of the obtained feature map is resized to 64$\times$64. The parameters $n$ and $K$ in crossed-mask generator are set to 3 and $\left\{ {2,4,8,16} \right\}$, respectively. The set of pooling ratios $R$ and $C_0 $ in restoration network are empirically set to $\left\{ {2,3,4,5} \right\}$ and 64.

During the training phase, our model is optimized by Adam optimizer with a learning rate of $1e^-4$, a weight decay of $1e^-3$ and a batch size of 6 for 400 epochs. The model is implemented by PyTorch and executed on a PC with an Intel i7-12700F CPU and an NVIDIA GeForce RTX 3090 GPU.

During the testing phase, the final anomaly map is an average of the anomaly maps generated by all $k$ values in the set $K$ respectively. And the individual anomaly map for each $k$ is computed by measuring the difference between the normalized input feature map $F\left( I \right)$ and the normalized reconstructed feature map $\hat F\left( I \right)$ using $L_2 (F(I),\hat F(I)) + (1 - SSIM(F(I),\hat F(I))) + (1 - GMS(F(I),\hat F(I)))$

\subsubsection{Competing methods}
We compare the proposed method against the following state-of-the-art unsupervised anomaly detection methods: 1) feature-based methods SPADE \cite{ref17}, Patch SVDD \cite{ref23}, GCPF \cite{ref3}, DifferNet \cite{ref30}, CS-Flow \cite{ref32}, MKDAD \cite{ref34} and STPM \cite{ref57}; as well as 2) reconstruction-based methods AE-ssim \cite{ref48}, GANomaly \cite{ref8}, OCR-GAN \cite{ref45}, DFR \cite{ref44}, UTRAD \cite{ref36}, RIAD \cite{ref41} and SSM \cite{ref43}. To ensure fair comparisons, we re-execute the released codes with recommended parameters of the above methods on our computation platform.

\subsection{Comparisons with State-of-the-arts}
\subsubsection{Detection Results on MVTec AD}
In Table \ref{table1}, the proposed MFRNet is compared with recent anomaly detection methods on each category of the MVTec AD dataset. As we can see, our method outperforms all other methods in terms of F1 in nine out of fifteen categories. Moreover, in the remaining six categories, our method also achieves the second-best performance, which is very close to the best. Among these six categories, DFR performs the best on the hazelnut, metal$\_$nut and pill, but is prone to significant performance degradation on the transistor and cable. Similarly, although GCPF yields the top one F1 on the cable, it is inferior to ours on the capsule and screw by a great gap. On the transistor category, the same situation occurred with the STPM. On the contrary, our approach produces more consistently excellent performance across all categories and achieves the highest mean F1 score 52.11$\%$. Therefore, it can be concluded that the proposed MFRNet is not limited to specific defects and is highly adaptable to different objects, shapes and texture variations.

\subsubsection{Detection Results on BTAD}
\begin{table}[]
	\centering
	\caption{Performance comparison of different methods on the BTAD dataset ($\%$).The top three performances are highlighted with \textbf{bold}, \underline{underline} and \uwave{wave line}}\label{table2}
	\setlength{\tabcolsep}{0.8mm}
	\renewcommand\arraystretch{1.5}
	\resizebox{\linewidth}{!}{ \begin{tabular}{l|c|c|c|c|c|c|c|c|c|c|c|c}
	\hline
	\multirow{2}{*}{Methods} & \multicolumn{4}{c|}{Product 01}        & \multicolumn{4}{c|}{Product 02}                & \multicolumn{4}{c}{Product 03}
	\\ \cline{2-13}
	& AUROC    & MAE     & ACC    & F1     & AUROC    & MAE     & ACC    & F1 & AUROC    & MAE     & ACC    & F1       \\ \hline
	Patch-SVDD  &76.13	&13.71	&85.14	&17.85	&43.98	&36.11	&60.97	&15.43	&77.53	&12.28	&86.89	&16.35   \\ \hline
	SPADE 		&82.84	&39.81	&60.03	&15.10	&81.17	&46.66	&54.04	&23.52	&81.46	&57.24	&43.56	&13.62   \\ \hline
	GCPF  		&77.87	&26.85	&72.76	&14.35	&\uwave{84.91}	&25.55	&72.60	&28.16	&82.76	&27.10	&71.60	&20.41   \\ \hline
	AE-ssim		&84.37	&14.57	&84.21	&21.77	&54.22	&19.92	&76.20	&14.89	&84.48	&11.38	&87.06	&18.56   \\ \hline
	GANomaly 	&85.74	&6.75	&90.60	&15.58	&56.16	&9.37	&86.61	&7.68	&76.09	&4.87	&94.35	&8.60	  \\ \hline
	STPM		&87.78	&23.13	&75.85	&28.12	&80.78	&34.08	&65.11	&\uwave{28.69}	&86.38	&34.11	&65.53	&21.74 \\ \hline
	MKDAD		&\underline{91.81}	&20.02	&77.46	&\uwave{42.06}	&75.82	&26.02	&71.70	&25.49	&\uwave{92.15}	&28.10	&71.14	&\uwave{34.21} \\ \hline
	DifferNet	&67.77	&\uwave{5.61}	&89.25	&7.33	&65.92	&\uwave{7.24}	&87.87	&6.89	&68.79	&\uwave{3.34}	&93.05	&5.58 \\ \hline
	CS-Flow		&78.14	&14.70	&81.22	&28.38	&63.26	&26.54	&73.13	&19.61	&73.16	&13.10	&85.47	&13.75 \\ \hline
	DFR			&\textbf{92.66}	&9.68	&89.76	&\textbf{45.01}	&\underline{90.44}	&16.94	&81.01	&\underline{30.50}	&\underline{93.62}	&15.21	&79.93	&\underline{41.93} \\ \hline
	OCR-GAN		&72.48	&6.88	&\textbf{95.62}	&3.56	&56.19	&\underline{7.06}	&\underline{94.88}	&2.56	&62.78	&4.25	&\uwave{97.57}	&1.22 \\ \hline
	UTRAD		&72.43	&\underline{5.50}	&\uwave{92.50}	&21.50	&57.16	&7.42	&\uwave{93.07}	&6.48	&58.24	&\underline{2.96}	&\underline{97.94}	&4.73 \\ \hline
	RIAD		&87.66	&21.66	&76.82	&26.30	&73.10	&23.30	&72.42	&27.15	&82.35	&18.54	&80.49	&13.54 \\ \hline
	SSM			&85.34	&15.28	&83.93	&19.76	&66.29	&8.20	&89.37	&13.62	&82.34	&10.22	&89.36	&12.70 \\ \hline
	Ours		&\uwave{91.52}	&\textbf{5.27}	&\underline{94.12}	&\underline{43.58}	&\textbf{92.45}	&\textbf{6.39}	&\textbf{95.41}	&\textbf{35.30}	&\textbf{94.34}	&\textbf{2.67}	&\textbf{98.88}	&\textbf{42.08} \\ \hline
\end{tabular}}
\end{table}
The quantitative comparison of different methods for each product of the BTAD dataset is listed in Table \ref{table2}. Since it is not very convincing to perform well on only one indicator, four metrics are used for a more comprehensive evaluation. It can be seen from Table \ref{table2} that for almost all four evaluation metrics, our method holds the best performance on each type of product. Specifically, for the product 01, our MFRNet ranks first in terms of MAE and can be in the top three in the other three metrics. Although DFR achieves slightly better AUROC and F1 scores, our approach outperforms it with respect to MAE and ACC scores by a large margin. And OCR-GAN has a higher ACC compared to ours, but its AUROC and F1 scores are extremely poor. Concerning the product 02 and product 03, our MFRNet consistently surpasses the other methods in terms of all four metrics. Remarkably, our approach far exceeds the second-best on the F1 score, obtaining a significant improvement of 15.7$\%$ and 0.36$\%$, respectively. The above results suggest that the proposed MFRNet has a more comprehensive performance compared to other existing methods.

\subsubsection{Detection Results on MT, MSD-US and Fabric-US}
\begin{table}[]
	\centering
	\caption{Performance comparison of different methods on the MT, MSD-US and Fabric-US datasets ($\%$).The top three performances are highlighted with \textbf{bold}, \underline{underline} and \uwave{wave line}}\label{table3}
	\setlength{\tabcolsep}{0.8mm}
	\renewcommand\arraystretch{1.5}
	\resizebox{\linewidth}{!}{ \begin{tabular}{l|c|c|c|c|c|c|c|c|c|c|c|c}
			\hline
			\multirow{2}{*}{Methods} & \multicolumn{4}{c|}{MT}        & \multicolumn{4}{c|}{MSD-US}                & \multicolumn{4}{c}{Fabric-US}
			\\ \cline{2-13}
			& AUROC    & MAE     & ACC    & F1     & AUROC    & MAE     & ACC    & F1 & AUROC    & MAE     & ACC    & F1       \\ \hline
		Patch-SVDD	&73.33	&28.61	&70.56	&25.23	&83.32	&12.97	&85.98	&13.16	&69.19	&34.30	&64.88	&11.52   \\ \hline
		SPADE		&69.82	&30.90	&68.13	&23.91	&64.78	&21.64	&77.29	&6.82	&85.07	&17.75	&80.67	&16.23   \\ \hline
		GCPF		&71.65	&24.81	&73.79	&23.19	&62.69	&16.04	&82.64	&7.47	&84.71	&12.68	&85.56	&16.90   \\ \hline
		AE-ssim		&79.49	&21.25	&80.63	&21.59	&90.01	&13.06	&86.35	&12.44	&87.78	&8.09	&89.27	&29.22   \\ \hline
		GANomaly	&61.69	&11.24	&87.83	&10.56	&83.41	&2.63	&89.94	&19.84	&84.48	&\textbf{1.15}	&\underline{94.73}	&26.10   \\ \hline
		STPM		&64.31	&35.55	&64.21	&21.63	&77.56	&28.88	&70.54	&7.14	&82.18	&20.91	&77.82	&15.82   \\ \hline
		MKDAD		&\uwave{84.28}	&23.42	&77.77	&\uwave{35.74}	&97.51	&13.01	&86.06	&32.90	&\uwave{88.22}	&12.45	&85.67	&30.56   \\ \hline
		DifferNet	&78.98	&\uwave{9.56}	&90.56	&17.09	&89.99	&\underline{2.16}	&\uwave{94.76}	&24.84	&77.65	&1.88	&90.28	&20.14   \\ \hline
		CS-Flow		&56.83	&28.97	&68.53	&21.63	&90.21	&11.89	&86.63	&12.33	&83.71	&8.18	&89.75	&17.67   \\ \hline
		DFR			&\textbf{89.14}	&12.49	&87.31	&\underline{40.32}	&\underline{98.05}	&7.53	&90.64	&\textbf{43.64}	&\underline{88.67}	&6.51	&90.36	&\underline{39.34}   \\ \hline
		OCR-GAN		&58.21	&9.72	&\uwave{91.25}	&3.74	&80.75	&\textbf{1.13}	&90.84	&19.69	&71.10	&\uwave{1.87}	&\uwave{91.68}	&7.96   \\ \hline
		UTRAD		&51.94	&\underline{8.69}	&\underline{98.19}	&1.37	&74.94	&4.80	&94.35	&5.84	&78.35	&14.38	&83.60	&22.59   \\ \hline
		RIAD		&73.40	&29.64	&69.74	&27.50	&\uwave{97.96}	&10.74	&87.91	&\uwave{39.72}	&85.79	&20.22	&76.68	&28.60   \\ \hline
		SSM			&76.82	&13.69	&86.50	&27.50	&97.26	&3.46	&\textbf{95.83}	&38.30	&84.23	&6.43	&88.48	&\uwave{37.54}   \\ \hline
		Ours		&\underline{88.86}	&\textbf{7.07}	&\textbf{98.75}	&\textbf{42.11}	&\textbf{98.32}	&\uwave{2.17}	&\underline{95.75}	&\underline{43.11}	&\textbf{88.78}	&\underline{1.80}	&\textbf{96.89}	&\textbf{39.66}   \\ \hline
	\end{tabular}}
\end{table}
Table \ref{table3} shows quantitative comparisons of different methods on the MT, MSD-US and Fabric-US datasets. On the MT dataset, it can be found that our detection method performs the best on the MAE, ACC and F1 scores and is only slightly inferior to the top-performing method DFR in terms of AUROC. Although DFR yields a marginally higher AUROC, the proposed MFRNet obtains remarkable improvements of 18.6$\%$ and 13.1$\%$ in terms of MAE and ACC respectively. Concerning the MSD-US dataset, our method surpasses other compared methods in terms of AUROC, with the highest score of 98.32$\%$. In addition, the MAE, ACC and F1 scores of our method are also of great impression, reaching 2.17$\%$, 95.75$\%$ and 43.11$\%$, respectively. For the Fabric-US dataset, our method markedly outperforms the other methods on most of the metrics, except MAE. Despite having the best MAE of GANomaly, it fails in the F1 metric. In general, as reported in Table \ref{table3}, our MFRNet is one of the top contenders in all evaluation metrics across the three industrial datasets, indicating that MFRNet can be an efficient solution for a variety of anomaly detection tasks.

\subsubsection{Inference speed}
\begin{table}[]
	\centering
	\caption{Running speed comparison of different methods (s)}\label{table4}
	\setlength{\tabcolsep}{1.2mm}
	\renewcommand\arraystretch{1.5}
	\begin{tabular}{l|c|c|c|c|c|c}
	\hline
	Methods    & MVTec AD & BTAD & MT   & MSD-US & Fabric-US & Mean \\ \hline
	Patch-SVDD & 0.36     & 0.45 & 1.66 & 0.15   & 0.14      & 0.55 \\ \hline
	SPADE      & 3.83     & 5.04 & 7.33 & 3.14   & 3.59      & 4.59 \\ \hline
	GCPF       & 1.15     & 1.73 & 2.65 & 1.15   & 1.34      & 1.60 \\ \hline
	AE-ssim    & 0.14     & 0.08 & 0.02 & 0.13   & 0.04      & 0.08 \\ \hline
	GANomaly   & 0.16     & 0.10 & 0.04 & 0.15   & 0.06      & 0.10 \\ \hline
	STPM       & 0.13     & 0.10 & 0.03 & 0.17   & 0.06      & 0.10 \\ \hline
	MKDAD      & 0.81     & 0.69 & 0.61 & 0.35   & 0.74      & 0.64 \\ \hline
	DifferNet  & 0.40     & 0.34 & 0.29 & 0.42   & 0.31      & 0.35 \\ \hline
	CS-Flow    & 0.17     & 0.12 & 0.07 & 0.17   & 0.09      & 0.12 \\ \hline
	DFR        & 0.16     & 0.12 & 0.03 & 0.41   & 0.06      & 0.16 \\ \hline
	OCR-GAN    & 0.23     & 0.23 & 0.08 & 0.26   & 0.14      & 0.19 \\ \hline
	UTRAD      & 0.10     & 0.10 & 0.04 & 0.12   & 0.06      & 0.08 \\ \hline
	RIAD       & 0.43     & 0.37 & 0.31 & 0.47   & 0.35      & 0.39 \\ \hline
	SSM        & 0.33     & 0.28 & 0.20 & 0.34   & 0.24      & 0.28 \\ \hline
	Ours       & 0.35     & 0.32 & 0.28 & 0.39   & 0.27      & 0.32 \\ \hline
\end{tabular}
\end{table}
We report the running time of different methods on five datasets for efficiency comparison. During the testing phase of MFRNet, multiple feature restorations are required to synthesize the detection results, which should be time-consuming. However, as listed in Table \ref{table4}, our method achieves an average inference time of 0.32s per image, which is in the middle of the existing methods, and acceptable for practical application requirements. We argue that our method can attain this considerably competitive speed, thanks to the light-weight structure of our restoration network. Meanwhile, parallel computation of multiple feature restorations or accelerated implementation with TensorRT can significantly reduce inference time. Furthermore, by utilizing smaller $k$ and $n$ in the crossed-mask generator, we can further obtain a more efficient detection method with only a small degradation in performance. The details can be seen in Section ‘Discussion’. In summary, the proposed method exhibits relatively high efficiency, while maintaining superior performance and strong applicability, making it a practical choice for real-world scenarios.

\subsection{Ablation Study}
To validate the effectiveness of each core component in our MFRNet, we conduct a series of ablation experiments on the bottle category of MVTec AD dataset. The first row in Table \ref{table5} indicates the baseline, which replaces the hybrid transformer block of restoration network with convolutional block and is trained directly on the normal image space using contextual loss $L_{Con} $. Based on this, we incrementally add the following two types of components: a) architecture: multi-scale feature aggregator (MFA), crossed-mask generator (CG) and hybrid transformer-based restoration network (HTR); b) loss: SSIM loss ($L_{SSIM} $) and GMS loss ($L_{GMS} $). The detailed results for each configuration are given below.

\subsubsection{Architecture ablation}
As listed in Table \ref{table5}, the baseline holds an unsatisfactory result and leaves room for improvement, due to the limited distinguishable information in image space and under-regularization of the model. Then, instead of reconstructing the raw image as in baseline, we attempt to reconstruct the corresponding feature map, which can provide more distinguishable representation. From the second row of Table \ref{table5}, we can observe that this modification is extremely beneficial, achieving a reduction of 10.7$\%$ in MAE, and contributing gains of 9.3$\%$, 7.6$\%$ and 37.5$\%$ in AUROC, ACC and F1 metrics respectively, compared to the baseline. We can also see that the reconstruction of the feature map can significantly suppress erroneously reported anomalies in Fig. \ref{fig6}(d). Next, the reconstruction problem is converted into a restoration problem by introducing CG module, and the convolutional network in baseline is still utilized to repair the masked feature map. We observe a substantial improvement in AUROC (86.68$\%$ v.s. 91.79$\%$), MAE (7.93 v.s. 7.65), ACC (85.73$\%$ v.s. 88.56$\%$) and F1 (42.85$\%$ v.s. 50.79$\%$). It can also be noted from Fig. \ref{fig6}(e) that the problem of anomalies being well reconstructed has been effectively addressed. Besides, when using the proposed restoration network with hybrid transformer blocks, the overall performance is further increased, showing it enjoys the benefits of both long-range contextual information and low-level details. A similar observation can be found in Fig. \ref{fig6}(f).
\begin{table}[]
	\centering
	\normalsize
	\caption{Ablation study of our MFRNet under different architectures and losses ($\%$).}\label{table5}
	\renewcommand\arraystretch{1.5}
	\resizebox{\linewidth}{!}{
		\begin{tabular}{ccc|ccc|c|c|c|c}
			\hline
			\multicolumn{3}{c|}{Architecture}                        & \multicolumn{3}{c|}{Loss}                                     & \multirow{2}{*}{AUROC} & \multirow{2}{*}{MAE} & \multirow{2}{*}{ACC} & \multirow{2}{*}{F1} \\ \cline{1-6}
			\multicolumn{1}{c|}{MFA} & \multicolumn{1}{c|}{CG} & HTR & \multicolumn{1}{c|}{$L_{Con} $} & \multicolumn{1}{c|}{$L_{SSIM} $} & $L_{GMS} $ &                        &                      &                      &                     \\ \hline
			\multicolumn{1}{c|}{}    & \multicolumn{1}{c|}{}   &     & \multicolumn{1}{c|}{$\checkmark$}     & \multicolumn{1}{c|}{}      &      & 79.31                  & 8.88                 & 79.70                & 31.16               \\ \hline
			\multicolumn{1}{c|}{$\checkmark$}    & \multicolumn{1}{c|}{}   &     & \multicolumn{1}{c|}{$\checkmark$}     & \multicolumn{1}{c|}{}      &      & 86.68                  & 7.93                 & 85.73                & 42.85               \\ \hline
			\multicolumn{1}{c|}{$\checkmark$}    & \multicolumn{1}{c|}{$\checkmark$}   &     & \multicolumn{1}{c|}{$\checkmark$}     & \multicolumn{1}{c|}{}      &      & 91.79                  & 7.65                 & 88.56                & 50.79               \\ \hline
			\multicolumn{1}{c|}{$\checkmark$}    & \multicolumn{1}{c|}{$\checkmark$}   &$\checkmark$     & \multicolumn{1}{c|}{$\checkmark$}     & \multicolumn{1}{c|}{}      &      & 95.54                  & 7.22                 & 90.07                & 57.01               \\ \hline
			\multicolumn{1}{c|}{$\checkmark$}    & \multicolumn{1}{c|}{$\checkmark$}   &$\checkmark$     & \multicolumn{1}{c|}{$\checkmark$}     & \multicolumn{1}{c|}{$\checkmark$}      &      & 96.86                  & 6.96                 & 90.80                & 61.33               \\ \hline
			\multicolumn{1}{c|}{$\checkmark$}    & \multicolumn{1}{c|}{$\checkmark$}   &$\checkmark$     & \multicolumn{1}{c|}{$\checkmark$}     & \multicolumn{1}{c|}{$\checkmark$}      &$\checkmark$      & 98.43                  & 6.88                 & 91.29                & 64.01               \\ \hline
	\end{tabular}}
\end{table}

\subsubsection{Loss ablation}
\begin{figure}[t]
	\centering
	\includegraphics[width=0.495\textwidth]{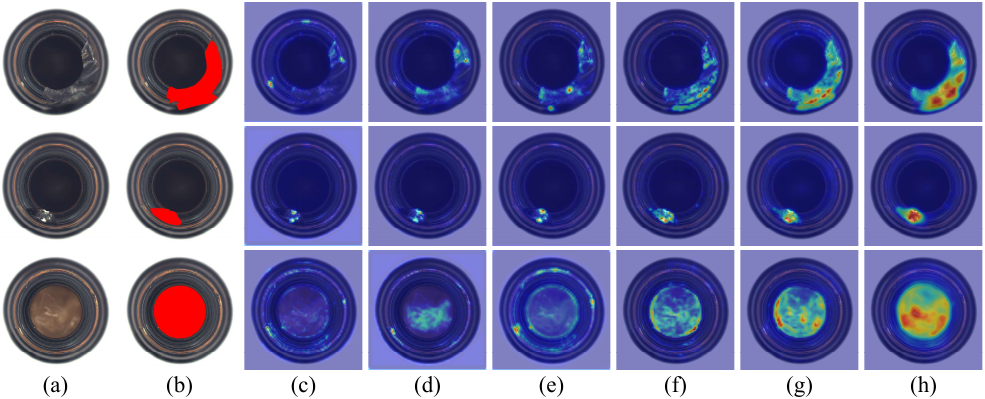}\\
	\caption{Visual comparison of our MFRNet under different architectures and losses. (a) Input image. (b) Input image overlaid with ground truth in red. (c)-(h) Anomaly score maps produced by each row of models in Table \ref{table5}, with low scores in blue and high scores in red.}\label{fig6}
\end{figure}

When a structural loss $L_{SSIM} $ is added to the training and testing of our entire architecture, the AUROC, ACC and F1 are respectively increased by 1.4$\%$, 0.8$\%$ and 7.6$\%$, while the MAE is reduced by 3.6$\%$. Finally, as can be seen in the last row of Table \ref{table5}, our MFRNet, equipped with hybrid loss, sets a new state-of-the-art and achieves 98.43$\%$ AUROC, 6.88$\%$ MAE, 91.29$\%$ ACC and 64.01$\%$ F1. Moreover, the visual comparison of our MFRNet under different losses is also presented in Fig. \ref{fig6} to signify the importance of the hybrid loss in a more intuitive way. It is clear that $L_{Con} $ can detect anomalous areas, but the defective areas are relatively discrete due to the per-pixel error measure. In contrast, the combination of $L_{Con} $ and $L_{SSIM} $ captures interdependencies between the local regions and produces more visually consistent detection results. In Fig. \ref{fig6}(h), the hybrid loss can highlight the complete defects with higher anomaly scores. The above results prove that all loss functions are essential for the proposed method to achieve the best anomaly detection results.

\section{Discussions}
In this section, we carry out a series of experiments to gain deeper insights into our MFRNet for unsupervised anomaly detection, all of which are conducted on the bottle category of MVTec AD dataset, unless otherwise stated.

\subsection{Influence of pre-trained model}
\begin{table}[]
	\centering
	\caption{Performance comparison of our method with different backbone networks ($\%$). Bold means the best.}\label{table6}
	\setlength{\tabcolsep}{2.5mm}
	\renewcommand\arraystretch{1.5}
	\begin{tabular}{l|c|c|c|c}
	\hline
	Backbone                   & AUROC & MAE   & ACC   & F1    \\ \hline
	AlexNet \cite{ref58}           & 93.40 & 9.44  & \textbf{91.41} & 59.86 \\ \hline
	ResNet18 \cite{ref59}          & 84.93 & 10.37 & 85.96 & 43.63 \\ \hline
	DenseNet201 \cite{ref60}       & 85.79 & 9.12  & 87.96 & 38.79 \\ \hline
	EfficientNet-B0  \cite{ref61} & 88.02 & 9.77  & 82.79 & 47.17 \\ \hline
	VGG16 \cite{ref50}             & \textbf{98.43} & \textbf{6.88}  & 91.29 & \textbf{64.01} \\ \hline
\end{tabular}
\end{table}
For the multi-scale feature aggregator in MFRNet, besides using VGG16 to obtain discriminant features, we also test our method with different backbone networks in Table \ref{table6}. These backbone networks are initialized with ImageNet pre-trained weights, and similarly, the first three feature blocks are chosen as the output. The size of input image and output feature is identical to that of the previous setting. It can be observed that the results are basically stable when using different backbones, suggesting that our method has a relatively strong robustness. However, we surprisingly find that models such as ResNet18, DenseNet201, and EfficientNet-B0 perform better on classification tasks but produce worse anomaly detection results. We speculate the underlying reason is that the features extracted by these models are so abstract that information about small anomalies is lost. Therefore, only choosing the earlier feature layers in these models may be a good solution. In summary, our MFRNet can be used as a plug-in model with arbitrary backbone networks, and we recommend using VGG16 in the multi-scale feature aggregator.

\subsection{Influence of multi-scale feature}
\begin{figure}[]
	\centering
	\includegraphics[width=0.495\textwidth]{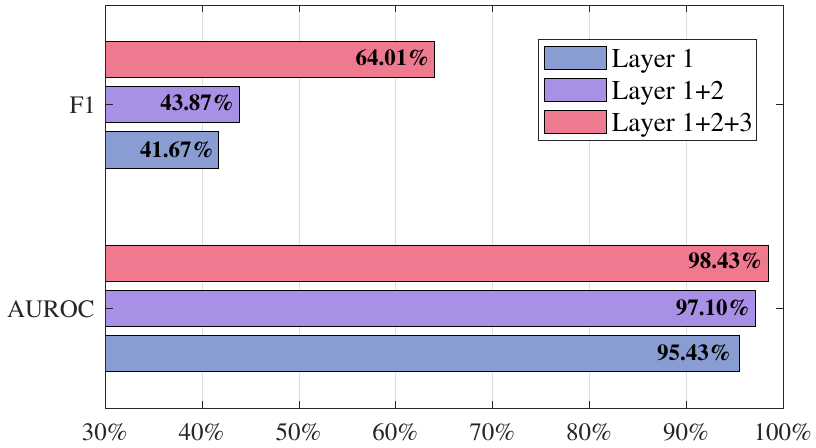}\\
	\caption{Performance comparison of our method with increasing feature layers.}\label{fig7}
\end{figure}

In the multi-scale feature aggregator, we extract hierarchical discriminative representations from different pre-trained feature layers. To quantify the effect of these multi-scale features, we show the results obtained when varying the feature layers extracted in VGG16. As illustrated in Fig. \ref{fig7}, the AUROC and F1 values of three configurations are reported, including using the first layer (Layer 1), the first two layers (Layer 1+2), and the first three layers (Layer 1+2+3). It is clear that features from the first layer can already yield encouraging results, and there is a consistent growth trend as the feature layers increase. This suggests that each feature layer can convey some different types of information that contribute to the final detection performance, and the performance may be further improved with more layers involved. Meanwhile, unlike previous work \cite{ref44} that simply concatenates features from different layers, our method can establish information interactions between different feature layers via a head convolution. This also keeps the computational cost of the method constant, regardless of the number of feature layers used. As a result, we extract the last feature layer in the first three blocks by default for the VGG16 in MFRNet. And ones can also set it up with the task complexity of the actual industrial scenario to make the most of its performance.

\subsection{Choice of masking size}
\begin{figure}[h]
	\centering
	\includegraphics[width=0.495\textwidth]{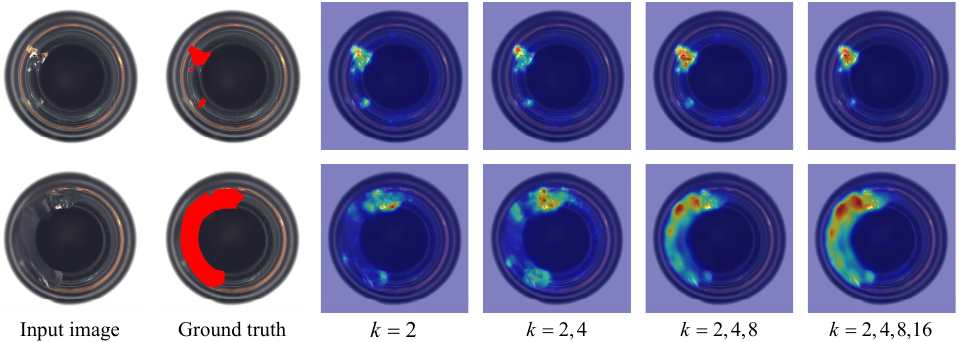}\\
	\caption{Anomaly maps produced by our method with different masking sizes. The first two columns are input image and input image overlaid with ground truth in red respectively. The remaining columns are anomaly score maps produced by our method with different masking sizes.}\label{fig8}
\end{figure}
Since anomalies in the real world vary in size, the crossed-mask generator should consider multiple masks with different sizes, and then the generated anomaly maps for each masking size should also be merged to compute the final anomaly map. To determine the choice of masking size in our method, we compare the performance when uniting different values of $k$. From Table \ref{table7}, going from a single masking size $k=2$ to multiple masking sizes $k{\rm{ = 2,4,8,16}}$, we observe an improvement of 1$\%$, 7.8$\%$, 1.7$\%$ and 35.7$\%$ in terms of AUROC, MAE, ACC and F1, showing that combining different masking sizes is beneficial. To provide an intuitive understanding of the effect of masking size in our method, we also illustrate the anomaly maps under different masking sizes in Fig. \ref{fig8}. As expected, the small-size mask helps to find the tiny anomalies, but struggles with the large ones. And when combined with more large-size masks, we can cover the anomalies with various sizes, thus generating more reliable anomaly maps. Therefore, in the latter experiments, a set of masking sizes $K \in \left\{ {{\rm{2,4,8,16}}} \right\}$ is utilized in the proposed method.

\begin{table}[t]
	\centering
	\caption{Performance comparison of our method with different masking sizes ($\%$).}\label{table7}
		\renewcommand\arraystretch{1.3}
		\resizebox{\linewidth}{!}{
		\begin{tabular}{cccc|c|c|c|c}
			\hline
			\multicolumn{4}{c|}{Masking size $k$}                                            & \multirow{2}{*}{AUROC} & \multirow{2}{*}{MAE} & \multirow{2}{*}{ACC} & \multirow{2}{*}{F1} \\ \cline{1-4}
			\multicolumn{1}{c|}{2} & \multicolumn{1}{c|}{4} & \multicolumn{1}{c|}{8} & 16 &                        &                      &                      &                     \\ \hline
			\multicolumn{1}{c|}{$\checkmark$}  & \multicolumn{1}{c|}{}  & \multicolumn{1}{c|}{}  &    & 97.44                  & 7.46                 & 89.76                & 47.18               \\ \hline
			\multicolumn{1}{c|}{$\checkmark$}  & \multicolumn{1}{c|}{$\checkmark$}  & \multicolumn{1}{c|}{}  &    & 97.99                  & 7.31                 & 90.74                & 52.99               \\ \hline
			\multicolumn{1}{c|}{$\checkmark$}  & \multicolumn{1}{c|}{$\checkmark$}  & \multicolumn{1}{c|}{$\checkmark$}  &    & 98.52                  & 7.05                 & 90.86                & 57.13               \\ \hline
			\multicolumn{1}{c|}{$\checkmark$}  & \multicolumn{1}{c|}{$\checkmark$}  & \multicolumn{1}{c|}{$\checkmark$}  & $\checkmark$   & 98.43                  & 6.88                 & 91.29                & 64.01               \\ \hline
	\end{tabular}}
\end{table}

\subsection{Choice of masking ratio}
\begin{figure}[htp]
	\centering
	\includegraphics[width=0.495\textwidth]{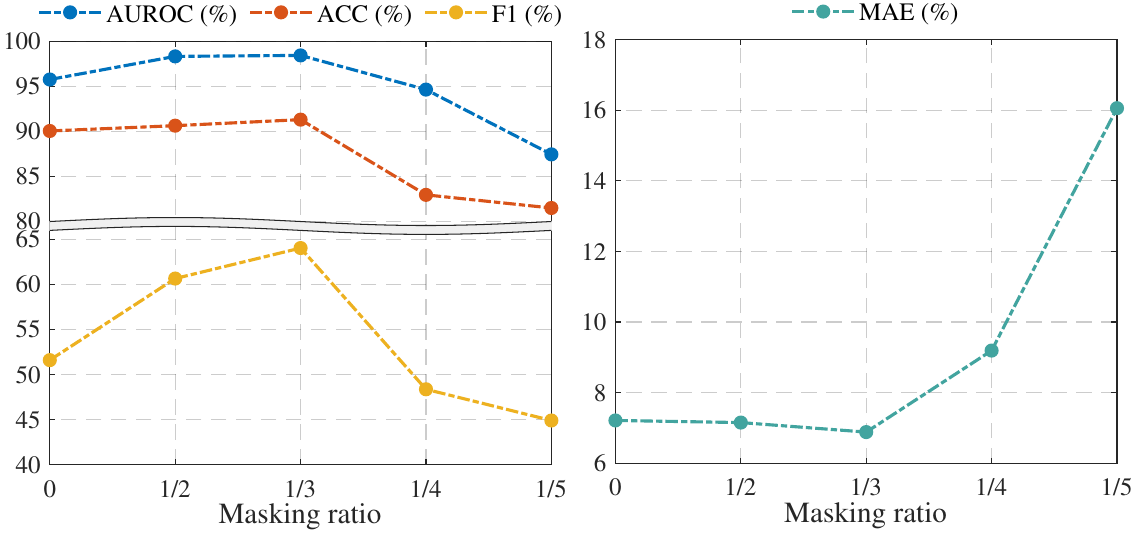}\\
	\caption{Performance comparison of our method with different masking ratios in terms of AUROC, ACC, F1 (left), and MAE (right).}\label{fig9}
\end{figure}
Besides the masking size discussed above, masking ratio is also an important parameter that affects the performance. The masking ratio refers to the ratio of patches removed in the feature map to be restored, which is equal to $1/n$. For example, in Fig. \ref{fig3}, $n=3$ means that the masking ratio is $1/3$, i.e., one-third of the patches are covered. To quantify its effect in our method, we compare the performance when varying the masking ratio. The results are shown in Fig. \ref{fig9}, where the symbol 0 on the x-axis indicates feature reconstruction without masking. We can see that the results regarding AUROC, ACC and F1 show a parabolic trend, rising first and then falling. More specifically, they increase steadily with the value of $n$, peaking at $n=3$, then gradually decline with the continuous increase of $n$. The MAE value shows the opposite trend and also reaches the optimal value at $n=3$. Accordingly, two conclusions can be drawn: 1) Converting the reconstruction problem to a restoration problems indeed improve detection performance. 2) A much high value of $n$ can adversely affect performance. We believe the reason for this is that a much high value of $n$ (i.e., a much low masking ratio) merely creates a simple task that can be easily solved by extrapolation from surrounding pixels, and thus fails to learn meaningful representations. Therefore, we set the masking ratio to $1/3$, i.e., $n=3$ in the crossed-mask generator for the best performance.

\subsection{Low-shot anomaly detection}
\begin{figure}[]
	\centering
	\includegraphics[width=0.495\textwidth]{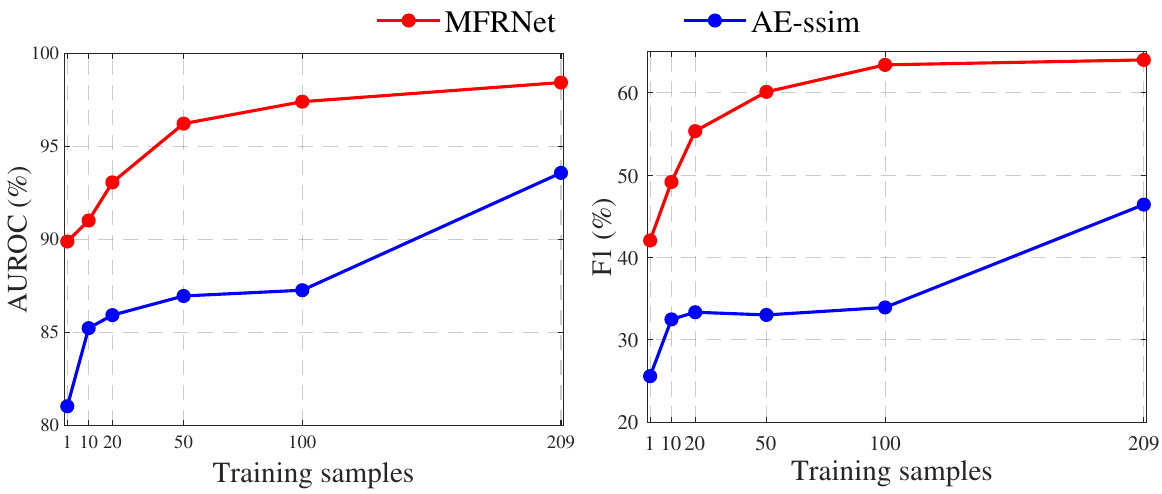}\\
	\caption{Performance comparison of our methods with different numbers of training samples.}\label{fig10}
\end{figure}
A better data efficiency is essential for real-world industrial inspections, where there may be a limited number of normal examples available. Therefore, we compare the performance of our method and AE-ssim \cite{ref48} with a smaller number of training samples, ranging from 1, 10, 20, 50, 100 and 209 (the full bottle category). As reported in Fig. \ref{fig10}, AE-ssim yields an apparent decrease of 6.7$\%$ and 26.9$\%$ in AUROC and F1 when reducing the training samples from 209 to 100, while our method yields only 1$\%$ and 0.98$\%$. And we also observe that our MFRNet with only 20 training samples can significantly outperform AE-ssim with a full 209 samples in terms of F1. Furthermore, it is worth noting that with only one training sample, our MFRNet still achieves noticeable performance and even is comparable to AE-ssim with full samples. We believe that the better data efficiency of our method is due to anomaly detection in feature space. Compared to the image space, our method can capture more distinguishable information even in the few-shot cases. Therefore, we argue that our method can be well applied in real-world industrial scenarios.

\subsection{Comparison with supervised methods}
In the above sections, we have verified that our MFRNet is better than the current unsupervised methods in terms of most evaluation metrics. And to further demonstrate its superiority, we also compare this unsupervised method with some supervised methods. Since supervised methods require anomaly labels during training, they are trained on the Fabric dataset \cite{ref56}, which contains 1200 abnormal fabric images with pixel-level annotations. As can be seen from Table \ref{table8}, it is surprising that our MFRNet achieves AUROC and ACC values close to the supervised methods, and even outperforms them in terms of F1 value. Regarding the MAE indicator, our method can perform better than most unsupervised methods, but worse than the supervised methods. We speculate the underlying reason is that there could be many plausible reconstructions for the same masked region, especially as the image pattern becomes more random and complex. In such cases, the plausible reconstruction of the masked area may differ significantly from the pixels or patches in the original image, thus resulting in increasing anomaly scores in these anomaly-free regions and a high MAE value of our method. In spite of this, considering that our MFRNet is trained in the unsupervised mode, we think that its performance is unparalleled and quite remarkable.
\begin{table}[]
	\centering
	\caption{Performance comparison of our method with supervised methods ($\%$).The top three performances are highlighted with \textbf{bold}, \underline{underline} and \uwave{wave line}.}\label{table8}
	\setlength{\tabcolsep}{2.7mm}
	\renewcommand\arraystretch{1.5}
	\begin{tabular}{l|c|c|c|c}
		\hline
		Methods       		& AUROC & MAE  & ACC   & F1    \\ \hline
		UNet \cite{ref62}  & \uwave{85.95} & \textbf{0.51} & \underline{97.15} & \underline{38.87} \\ \hline
		MedT \cite{ref63}  & 57.74 & \uwave{0.83} & 95.55 & 24.97 \\ \hline
		SETR \cite{ref64}  & \textbf{89.32} & 1.60 & 96.80 & \uwave{33.17} \\ \hline
		MCnet \cite{ref65} & 76.33 & \underline{0.59} & \textbf{97.75} & 29.73 \\ \hline
		Ours           		& \underline{88.78} & 1.80 & \uwave{96.89} & \textbf{39.66} \\ \hline
	\end{tabular}
\end{table}

\subsection{Visual presentation}
\begin{figure*}[]
	\centering
	\includegraphics[width=0.98\textwidth]{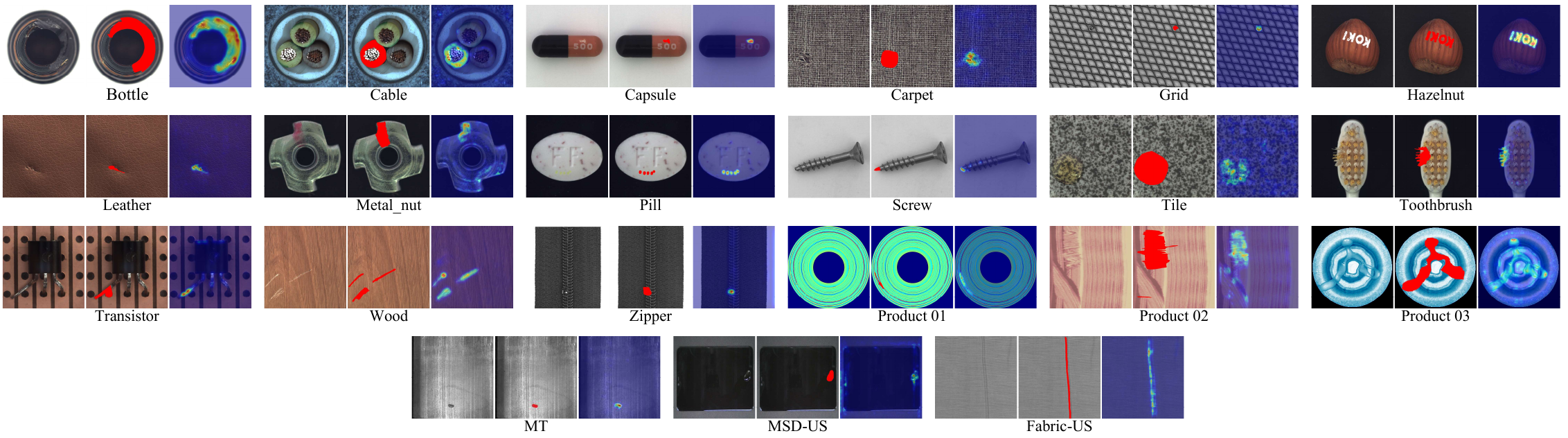}\\
	\caption{Anomaly detection results of our method. For each type of anomaly, the three columns from left to right are respectively input image, input image overlaid with ground truth in red, and anomaly detection map.}\label{fig11}
\end{figure*}
Some qualitative results of our anomaly detection method are presented in Fig. \ref{fig11}. As we can observe, the proposed method can accurately locate defects on different industrial material surfaces, such as fabric, wood, leather, metal, etc. Although the anomalies differ in size, shape, color and brightness, our method works well and thus is highly applicable to production lines. For example, our method can inspect the abnormal colors on the cable and metal$\_$nut categories with high accuracy and even recognize the misprinted words on the hazelnut category. Meanwhile, a tiny missing tip in the screw category is also precisely detected. In the tile category and MSD-US dataset, the abnormal oils appear transparent and have low visibility against the background, yet our method still detects them successfully, showing a strong semantic capability. Even on some material surfaces with complex texture variations, such as carpet, wood categories and Fabric-US dataset, our method can assign high anomaly scores to the anomalous regions.

\section{Conclusion and future work}
This paper proposes a multi-feature reconstruction network, MFRNet, based on crossed-mask restoration for unsupervised anomaly detection, which can detect anomalies accurately only using anomaly-free images. A multi-scale feature aggregator is adopted to extract discriminative hierarchical features of the raw image from a pre-trained model. A crossed-mask restoration network is also proposed to partially cover the generated multi-scale representation, and then restore the missing regions. Combined with the hybrid loss, MFRNet is able to ensure feature reconstruction and anomaly detection from both the pixel and structural similarity. Experimental results on four public datasets and our newly-proposed dataset show that our method is at par with or outperforms other state-of-the-art methods. In addition, our method has strong robustness and generalization ability and can even deliver detection performance close to the supervised methods. 

In the future, we plan to design a more effective and reasonable masking strategy to formulate the restoration task. Moreover, developing a stronger restoration network could be considered to further improve the detection accuracy and inference speed.

\vfill

\end{document}